\documentclass[letterpaper, 10 pt, conference]{ieeeconf}  

\IEEEoverridecommandlockouts                              

\usepackage{amsmath}

\usepackage[separate-uncertainty=true, multi-part-units=single, product-units=single]{siunitx}
\usepackage{graphicx}

\usepackage{graphics} 
\usepackage{epsfig} 
\usepackage{mathptmx} 
\usepackage{times} 
\usepackage{amsmath} 
\usepackage{amssymb}  

\title{\LARGE \bf
Real-time Deformation-aware Control for Autonomous Robotic Subretinal Injection under iOCT Guidance}

\author{Demir Arikan$^{1,2}$, Peiyao Zhang$^{2}$, Michael Sommersperger$^{1}$, Shervin Dehghani$^{1}$, Mojtaba Esfandiari$^{2}$,\\Russel H. Taylor$^{2}$,  M. Ali Nasseri$^{1,4}$, Peter Gehlbach$^{3}$, Nassir Navab$^{5}$ and Iulian Iordachita$^{2}$%
\thanks{
*This work is supported by the U.S. National Institutes of Health under the grants number R01EB023943, R01EB025883, R01EB34397, and partially by JHU internal funds.}%
\thanks{$^{1}$D. Arikan, M. Sommersperger, S. Dehghani, M. Ali Nasseri are with Department of Computer Science, Technische Universit{\"a}t M{\"u}nchen, Munich 85748 Germany {\tt\small demir.arikan@tum.de}}%
\thanks{$^{2}$D. Arikan, R. H. Taylor and I. Iordachita are with Laboratory for Computational Sensing and Robotics, Johns Hopkins University, Baltimore, MD, USA}%
\thanks{$^{3}$P. Gehlbach is with Wilmer Eye Institute, Johns Hopkins Hospital, Baltimore, MD, USA}%
\thanks{$^{4}$M. Ali Nasseri is with Augenklinik und Poliklinik, Klinikum rechts der Isar der Technische Universit{\"a}t M{\"u}nchen, M{\"u}nchen 81675 Germany}
\thanks{$^{5}$N. Navab is a full professor and head of the Chair for Computer Aided Medical Procedures \& Augmented Reality, Technical University of Munich, 85748 Munich, Germany}
}

\begin{document}

\maketitle
\thispagestyle{empty}
\pagestyle{empty}

\begin{abstract}

Robotic platforms provide consistent and precise tool positioning that significantly enhances retinal microsurgery. Integrating such systems with intraoperative optical coherence tomography (iOCT) enables image-guided robotic interventions, allowing autonomous performance of advanced treatments, such as injecting therapeutic agents into the subretinal space.
However, tissue deformations due to tool-tissue interactions constitute a significant challenge in autonomous iOCT-guided robotic subretinal injections.
Such interactions impact correct needle positioning and procedure outcomes.
This paper presents a novel method for autonomous subretinal injection under iOCT guidance that considers tissue deformations during the insertion procedure. 
The technique is achieved through real-time segmentation and 3D reconstruction of the surgical scene from densely sampled iOCT B-scans, which we refer to as B${^5}$-scans. Using B${^5}$-scans we monitor the position of the instrument relative to a virtual target layer between the ILM and RPE.
Our experiments on \textit{ex-vivo} porcine eyes demonstrate dynamic adjustment of the insertion depth and overall improved accuracy in needle positioning compared to prior autonomous insertion approaches.
Compared to a 35\% success rate in subretinal bleb generation with previous approaches, our method reliably created subretinal blebs in 90\% our experiments.
The source code and data used in this study are publicly available on GitHub\footnote{https://github.com/demirarikan/virtual-layer-retinal-surgery}.

\end{abstract}

\section{INTRODUCTION}

Subretinal injection is a surgical technique to deliver therapeutic agents, such as gene therapy vectors, transplanted cells or stems cells directly under the retina, specifically targeting genetic retinal dystrophies and degenerations. Such procedures may enable viable treatment options for prevalent diseases such as age-related macular degeneration (AMD), which alone is expected to affect 288 million people by 2040 \cite{wong2014global}.
Compared to injections into the vitreous with current anti-VEGF \cite{finger2020anti} drugs, direct injection of the payload into the subretinal space has the potential to maximize the therapeutic efficacy and reduce the systemic toxicity potential \cite{subretinal-irigoyen}.

\begin{figure}
    \centering
    \includegraphics[width=1\linewidth]{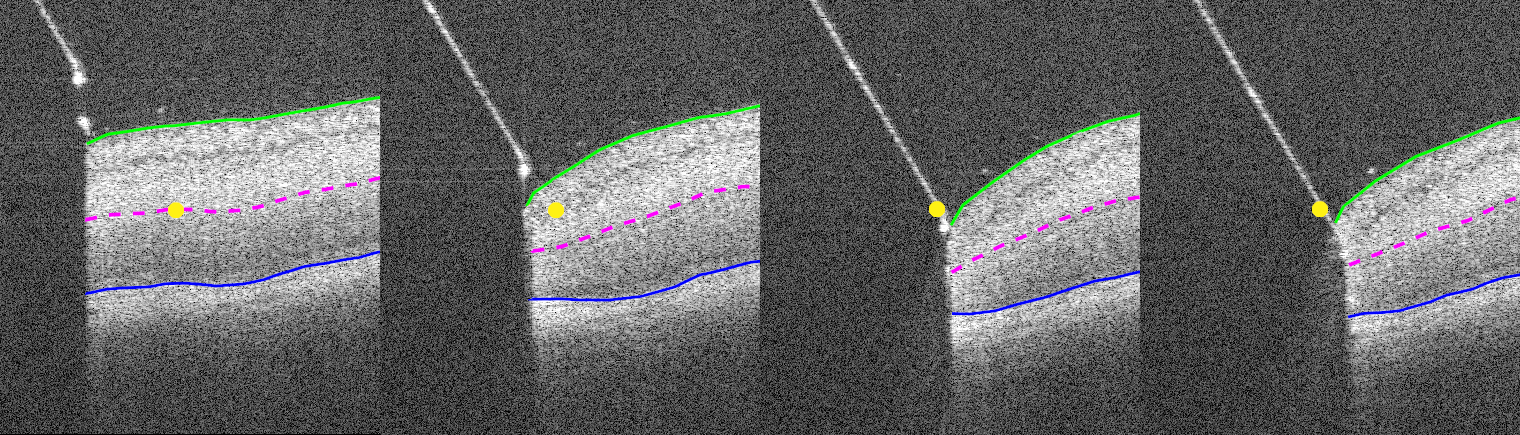}
    \caption{Tissue deformation during subretinal injection with a target (pink dashed line), defined as a virtual layer between ILM (green) and RPE (blue). In comparison, the fixed target point (yellow) is outside of the retina at the end of the insertion.}
    \label{fig:rel-vs-fix}
\end{figure}

Subretinal injections are performed by inserting a microsurgical cannula transretinal through the internal limiting membrane (ILM), and slowly guiding the tip to a point located between the photoreceptor cells and the retinal pigment epithelium (RPE)-Bruch's membrane complex. For optimal therapeutic results, a subretinal bleb, which can measure several millimeters in diameter, allows the therapeutic agent to reach the targeted subretinal space.

Due to the fragile and non-regenerative nature of the photoreceptor and RPE cells, any tissue damage can cause irreversible harm to the patient's eyesight \cite{George2021-ft}. 
Therefore, it is crucial to precisely place the needle at an optimal depth in the, on average 250 \textmu m thick \cite{ret-thickness}, human retina to minimize trauma while maximizing drug-tissue interaction.

There are two major difficulties in repeatedly performing successful subretinal injection: (1) visualization of the anatomical target area and (2) precise needle insertion without causing tissue damage.
The conventional way to visualize ophthalmic procedures is through an operating microscope.
In addition to the microscopic view, optical coherence tomography (OCT) can provide high-resolution cross-sectional imaging at high update rates to visualize the retinal layers and monitor the subretinal injection procedure.
OCT enables depth-resolved imaging, acquiring 2D B-scans and 3D C-scans of the surgical site \cite{Britten23, Li23, Vajzovic2019}.
\begin{figure*}[t]
    \centering
    \includegraphics[trim={0 0 0cm 0}, clip,width=0.93\linewidth]{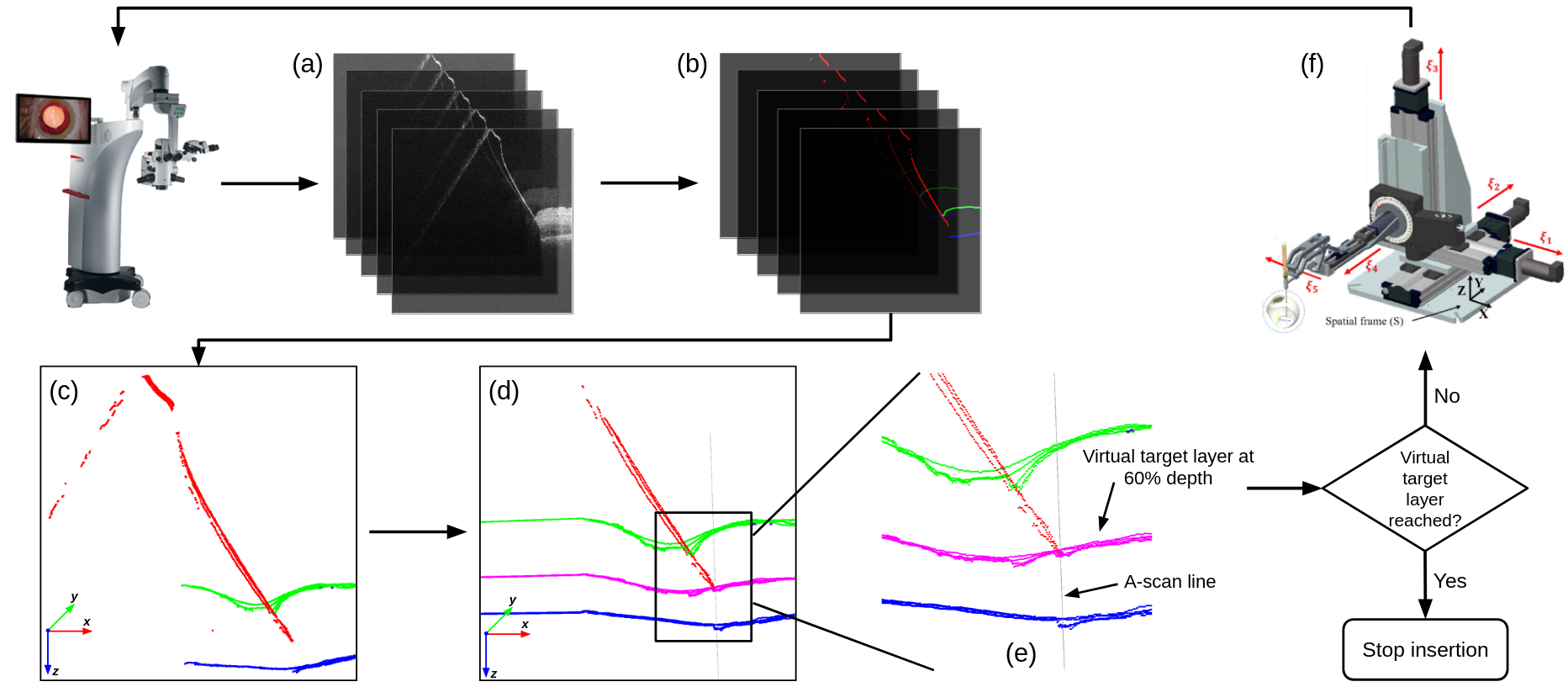}
    \caption{Overview of our real-time needle insertion pipeline.
    (a) B$^{5}$-scan, consisting of five 2D B-scans,  showing different cross sections of the needle and retina.
    (b) Segmented B-scans using our segmentation network, needle top surface (red), ILM (green), RPE (blue).
    (c) Point cloud generated from segmentation results without processing.
    (d) Point cloud generated from segmentation results after processing with inpainted ILM and RPE layers, removed outlier needle points and visualized virtual target layer (pink).
    (e) Needle tip area of the processed point cloud. Pink points are the virtual target layer, gray line represents the A-scan going through the needle tip.
    (f) Robot control adjustments based on needle position.
    }
    \label{fig:method-steps}
\end{figure*}
The second major challenge is guiding the surgical cannula to the target location.
Due to human physiological limits, such as the tremor of the surgeon's hand, which is around 180 \textmu m \cite{riviere_study_2000}, precise needle insertion and positioning at the required micrometer accuracy is naturally challenging. 
On the other hand, robotic systems \cite{vander-poorten, Molaei2017, Nasseri2013, He2012} have been developed for ophthalmic applications, which are not affected by limitations in human dexterity and enable precise and repeatable needle control inside the eye.
Paired with iOCT imaging, such robotic systems have been successfully employed for subretinal injections \cite{Yang2022-vf, Ladha2021, nasseri2017}. 
With advances in deep learning, OCT-guided robotic task autonomy has been enabled for subretinal injections \cite{Dehghani-23,Kim-23, Zhang2024-bt}.
While these approaches show promising results on \textit{ex-vivo} porcine eye experiments by navigating the cannula along a calculated trajectory to a target location, dynamic tissue behavior during the insertion procedure has not yet been fully considered.
Recent studies have demonstrated the effects of tool-tissue interaction during subretinal injection and captured the resulting dynamic changes in the tissue structures by OCT imaging \cite{pannek}.
Hence, real-time feedback mechanisms need to be developed to continuously monitor the needle insertion, update the control strategy according to tool-tissue interactions, and ensure correct targeting.

To address this challenge, we propose a real-time, OCT-based feedback system.
Our method utilizes OCT B$^{5}$-scans, which we define as five densely sampled 2D OCT B-scans combined together to generate a small 3D OCT volumes. 
B$^{5}$-scans are smaller than conventional C-scans and can, therefore, be updated at a high frequency, enabling real-time acquisition and processing that determines the robotic control strategy.
Retinal deformations, because of tool-tissue interactions, cause the retinal tissue, and therefore the target point, to shift.
Tracking the target point is a challenging computer vision task due to the retina's monotonous nature.
Meanwhile, the target point drifts in the $X$ and $Y$ directions are less critical because of the large size of the bleb; however, maintaining accurate positioning along the $Z$-axis is crucial, as it ensures that injections consistently reach the intended tissue layer, with minimal tissue injury.

Thus, we define the target area as a virtual layer located at a relative depth level between the retina's ILM and RPE layers rather than a fixed point in the 3D volume. 
This dynamic targeting method adapts to tissue deformation during the insertion procedure by continuously updating the target distance based on the current retinal conformation. This ensures successful needle insertion to the anatomical target, as visualized in Fig.~\ref{fig:rel-vs-fix}.

Overall, the main contributions of this work are as follows:
\begin{enumerate}
    \item We develop and evaluate a real-time method for dynamically controlling needle insertions under iOCT guidance during robotic subretinal injections.
    \item We introduce a virtual target layer located at a relative level in between ILM and RPE that dynamically adapts to tissue deformations caused by the needle insertion procedure.
\end{enumerate}
To validate our proposed method, we conducted two experiments on \textit{ex-vivo} open-sky porcine eyes. 
Additionally, we compared the accuracy and quality of subretinal injections using our virtual target layer method with the conventional fixed-point targeting approaches used in prior work utilizing a syringe pump.

\section{RELATED WORKS}
Our method is based on positioning the needle on a virtual target layer defined at a relative position between the ILM and RPE to compensate for tissue deformations during the insertion.
In the past, real-time 4D OCT-based distance estimation between surgical tools and the ILM and RPE surface has been investigated by Sommersperger \textit{et al.} \cite{Sommersperger:21}.
They estimated the tooltip position inside OCT volumes from 2D OCT B-scans extracted from the region around the surgical tool \cite{Weiss2020-fm}. 
They further segmented the B-scans and converted the segmentation maps into a 3D surface point cloud to calculate the distance between the tooltip and ILM and RPE layers. 
While this work presents an efficient real-time processing pipeline, the pipeline was not integrated into a robotic system or combined with a robot control strategy.

To advance robotic navigation autonomy for subretinal injection, Dehghani et al. \cite{Dehghani-23} used iOCT volumes to select a target location within the retina and plan the robot's trajectory.
To reduce the computations, they identified and segmented the instrument-aligned volume cross-section and estimated the needle pose.
Based on the instrument and target information, autonomous needle insertion was performed by aligning the needle with a planned insertion line and advancing the robot in the insertion direction.
In experiments on \textit{ex-vivo} open-sky porcine eyes, they achieved an average Euclidian error of $32 \pm 4$ \textmu m between the target point and the needle tip.

Kim et al. \cite{Kim-23} built a deep learning based framework for autonomous needle navigation in subretinal injections, where the needle is guided using visual feedback from both top-down microscope images and B-scans. 
After the needle tip was aligned with the user-defined target in the microscope image and touched the ILM layer in the B-scan, the insertion was done by selecting the target position in a 2D B-scan image and inserting it along its needle axis. Similarly to \cite{Dehghani-23}, they achieved a mean Euclidean error of 26 \textmu m between the target and needle tip and a depth error of $7 \pm 11$ \textmu m.

These studies evaluated their errors by comparing the final position of the needle tip in OCT scans with the intended target point. However, they only confirmed whether the correct spatial location was reached, even though the anatomical target may have shifted from that location.

The compression of the retina during insertion causes the tissue around the target to shift, leading to a corresponding displacement of the target itself.
Therefore, to overcome this limitation, we propose a novel dynamic deformation-aware control strategy based on a virtual target layer defined at a relative distance between the ILM and RPE, ensuring the needle is placed in the intended anatomical area. 

\section{METHOD}
Our deformation-aware robotic control strategy is enabled by real-time OCT volume processing.
An overview of the method is depicted in Fig.~\ref{fig:method-steps}.
Like \cite{Sommersperger:21}, the acquired OCT data is segmented and converted into point clouds of the ILM, RPE, and needle surface. 
The 3D point clouds are then processed to update the virtual target layer and estimate the needle tip position in relation to it.
During this process, we also consider typical OCT artifacts, such as gaps in the retinal layers induced by surgical tools with highly reflective metallic surfaces, which obstruct the OCT signal from reaching the underlying retina. 
An inpainting method is, therefore, applied to the point cloud to achieve a more accurate segmentation of the ILM and RPE layers.
With the segmentations of the tool, ILM, and RPE available, the position of the tooltip regarding the virtual target layer is calculated, and the robot is controlled accordingly.

\subsection{B$^{5}$-scans}
Currently, 4D OCT-capable systems are not yet commercially available.
To achieve real-time image acquisition without having an MHz swept-source OCT system, we propose to acquire only a few OCT B-scans that can be combined into a volume.
For reference, using our setup (Leica Proveo 8 with EnFocus OCT imaging system), we can acquire volumes at an imaging area of \qtyproduct{2x5}{\milli\meter}. The volumes consists of 200 B-scans, each composed of 1000 A-scans with 1024 pixels depth resolution and are acquired at a frame rate of 0.16 Hz (6000ms/volume), which does not allow for interactive update rates.

Our method uses B$^{5}$-scans as a compromise between conventional C-scans and single 2D B-scans.
We define B$^{5}$-scans as volume scans that consists of five equally spaced 2D B-scan images (Fig. \ref{fig:method-steps}(a)), covering a small scanning area. 
By reducing the scan area to \qtyproduct{4x0.1}{\milli\meter} and the number of B-scans per volume to five, a frame rate of approximately 9 Hz (115ms/volume) can be achieved with our setup.
The size and number of B-scans are carefully selected to balance the trade-offs between minimizing scan duration and ensuring the needle remains within the imaging area. 

\subsection{Dynamic Virtual Target Layer}
For our adaptive deformation-aware control strategy, we define the target not as a fixed 3D location in space but as a dynamic virtual target layer $L_{\text{target}}$ located at a relative level in between ILM and RPE.

Given a target point $i=(x, y, z)$ in 3D OCT volume coordinates, we define the constant $p$ as:
\begin{equation}
    p = \frac{z-d_{\text{ilm}}(x, y)}{d_{\text{rpe}}(x, y)-d_{\text{ilm}}(x, y)}
\end{equation}

where $d_{c}(x, y)$ denotes the pixel index of the first occurrence of class $c \in \{\text{ilm}, \text{rpe}, \text{tooltip}, \text{target point}\}$ along the A-scan located on $(x, y)$ coordinates.

Thereafter, we define the virtual layer $L_{\text{target}}$ as:

$$
L_{\text{target}}(x, y) = d_{\text{ilm}}(x, y) + p \cdot \left( d_{\text{rpe}}(x, y) -d_{\text{ilm}}(x, y) \right) 
$$

for all the $(x, y)$ pairs in the B$^{5}$-scan, continuously at each time frame. 
This definition of the virtual target layer is based on the anatomical structure of a healthy retina. However, the presence of blebs, retinal detachments, or retinal dystrophies may alter the injection site and technique.

By determining the $p$ value of the target in the initial B$^{5}$-scan and calculating $L_{\text{target}}$ in each subsequent B$^{5}$-scan, we redefine the target from a fixed point to a surface $L$, ensuring that any axial target movement is accounted for.
All the points on $L_{\text{target}}$ rely on the positions of the retinal layers and, therefore, directly compensate tool-tissue interactions that result in tissue deformation and compression of the retina.

\subsection{OCT Segmentation Network}
\begin{figure}
    \centering
    \includegraphics[trim={0 10cm 0 0}, clip, width=\linewidth]{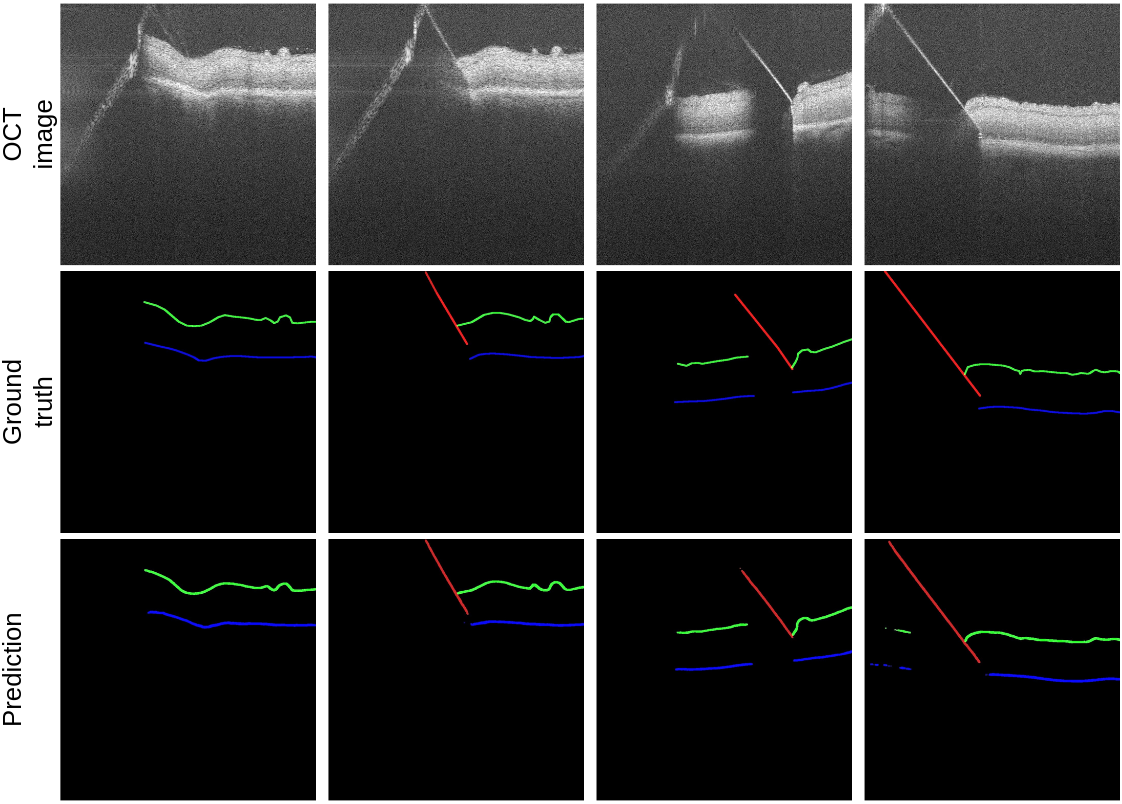}
    \caption{Example input and ground truth pairs from our dataset. Needle top surface (red), ILM (green), RPE (blue).}
    \label{fig:network-example}
\end{figure}

To calculate the needle position regarding the virtual target layer, we segment the needle, ILM, and RPE surfaces in the B-scans after each B$^{5}$-scan acquisition. Like \cite{Sommersperger:21}, we train a U-Net \cite{ronneberger2015u} segmentation network with Dice loss and Adam optimizer \cite{Kingma2014AdamAM} using the MONAI framework \cite{cardoso2022monaiopensourceframeworkdeep}. 
The segmentation network was trained on a custom dataset of 1707 2D OCT images collected from B$^{5}$-scans of robotic subretinal injections. 
The data was collected using the setup described in section \ref{experimental-setup}. 
The images were labeled by a biomedical engineering expert using the SUPERVISELY\footnote{https://supervisely.com/} image labeling platform. 
Fig. \ref{fig:network-example} shows example input and ground truth images from our dataset.

\subsection{Point Cloud Processing}
Similar to \cite{Sommersperger:21}, we convert the segmentation results of all B-scan slices into 3D point clouds by selecting the first occurrence of each class along every A-scan. 
The coordinates of each class pixel correspond to its position inside the 3D point cloud. 
To remove mislabeled needle points, we fit a straight line, which is the same shape as our needle, using RANSAC \cite{ransac} to the needle points. We remove needle points that lie outside of a distance threshold. 
To reconstruct areas of the retinal layers shadowed by the metallic needle we use the retinal inpainting method described in \cite{Sommersperger:21}. 

\begin{figure}
    \centering
    \includegraphics[trim={0 0 0 0}, clip, width=0.9\linewidth]{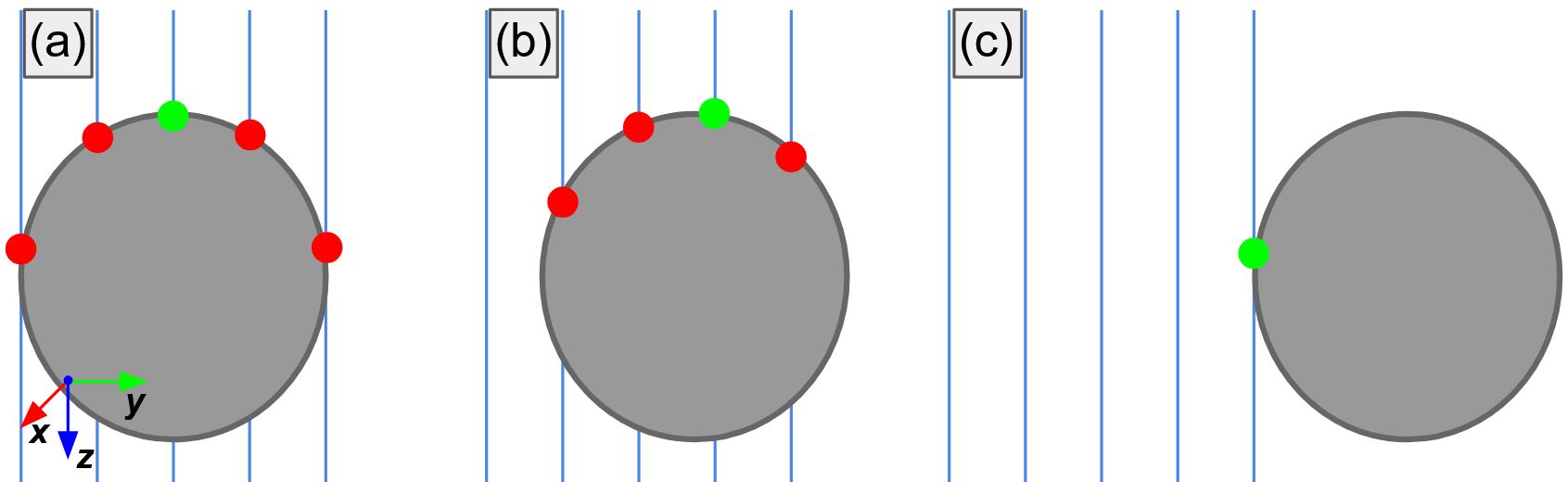}
    \caption{Front view diagram of the needle (gray), scan lines (blue) and their intersections corresponding to points in the point cloud (red and green). (a) Perfect alignment of scan lines and needle, parts of the needle visible in each scan line. (b) Average alignment, few scan lines do not contain needle. (c) Worst case alignment. In each case the point closest to the center of the needle higher than others (green). Worst case needle tip depth error is half needle thickness.}
    \label{fig:needle-tip-cases}
\end{figure}

We define the needle tip as one of the tip points in the point cloud closest to the needle's center. Due to its circular shape, 
the point closest to the center is also above all other points. Hence, we select the highest point to be the needle tip (green points in Fig. \ref{fig:needle-tip-cases}). 

\subsection{Robot Control}
The alignment of the robot and the OCT microscope is essential for successful robotic targeting. 
We use the setup presented in \cite{Dehghani-23}, positioning the robot in proximity to the retina and aligning the B$^5$-scans with the needle.
To minimize interaction with the retina, the needle is only moved along its insertion direction and angled 45 degrees to the retinal surface.
By estimating the needle tip position and target area, the state of the needle tip, whether it is within the target area or not, and the needle tip distance to the virtual target layer is assessed following each B$^{5}$-scan acquisition.
To guarantee a precise, yet efficient targeting, the velocity of the insertion is adjusted while the needle is continuously moved deeper inside the tissue using the following equation:

\begin{equation}
  v_{c}=\left\{
    \begin{array}{ll}
      v_{m}, & \text{if}\; d_{\text{tooltip}} < d_{\text{ilm}}\\

      |\cfrac{\text{dist}(d_{\text{tooltip}}, L_{\text{target}})}{d_{\text{rpe}} - d_{\text{ilm}}}| \cdot  v_{m}, & \text{if}\;  |\cfrac{\text{dist}(d_{\text{tooltip}}, L_{\text{target}})}{d_{\text{rpe}} - d_{\text{ilm}}}| > \alpha\\

      0 & ,\text{otherwise} \\
            
    \end{array}
  \right.
\end{equation}

Here, $v_{c}$ and $v_{m}$ denote the current and predefined maximum velocities and $d_{.}$ is shorthand form of $d_{.}(x, y)$, where $(x, y)$ is the location of the A-scan on which the needle tip is located. The factor $\alpha \in [0, 1]$ controls the stopping criterion. 

\section{EXPERIMENTS} \label{experimental-setup}
\begin{figure}
    \centering
    \includegraphics[trim={0 0 0 2cm}, clip, width=0.9\linewidth]{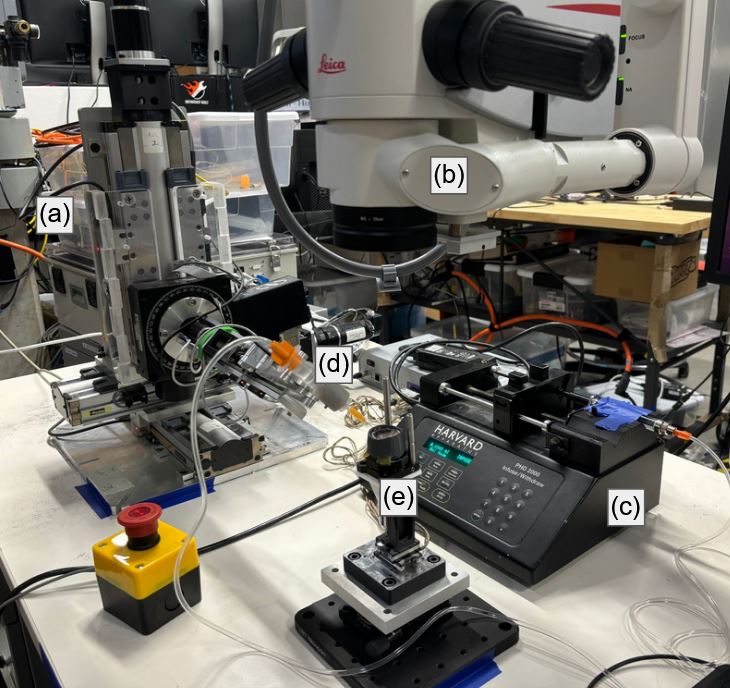}
    \caption{Overview of our experimental setup. (a) Steady Hand Eye Robot \cite{He2012}, (b) OCT imaging system (Proveo 8 with EnFocus, Leica, Germany), (c) Syringe pump (PHD2000, Harvard Apparatus, USA), (d) Syringe with 42 gauge needle (INCYTO Co., Ltd., South Korea), (e) \textit{Ex-vivo} open-sky porcine eye on 3D printed container.}
    \label{fig:experimental-setup}
\end{figure}
An overview of our experimental setup and equipment can be seen in Fig. \ref{fig:experimental-setup}. Open-sky refers to the removal of the anterior eye segment, providing easier access to the retina. 

\begin{figure}[h]
    \centering
    \includegraphics[width=0.9\linewidth]{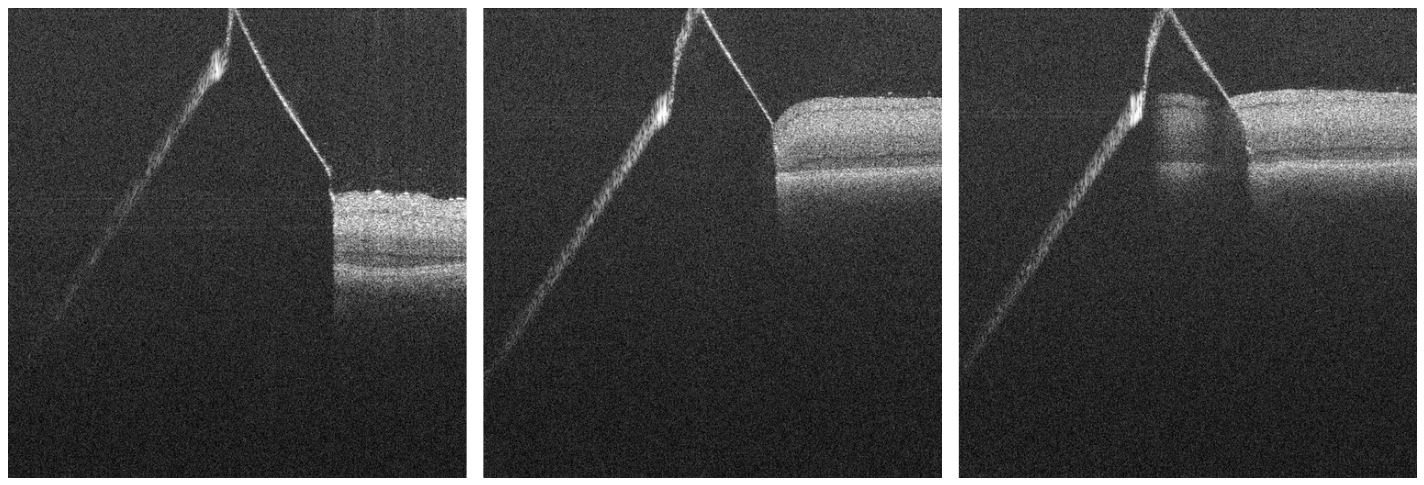}
    \caption{Example OCT images of the three phases during a needle insertion. Outside (left), border (middle) and inside (right). }
    \label{fig:phases-example}
\end{figure}

For the first experiment, we evaluated the accuracy of our methods in estimating needle tip, ILM and RPE positions along the needle tip A-scan using previously collected OCT volumes. These volumes were categorized into three phases (Fig. \ref{fig:phases-example}), \textbf{Outside}: The needle is completely outside the retina, \textbf{Border}: The needle is pressing against retina or tip slightly inside, causing significant deformation of the ILM, \textbf{Inside}: The needle is inside the tissue with no or minimal tissue deformation.
We utilized ten volumes each for the outside and border phases and twelve for the inside phase.
For the second experiment, we compared our method with a fixed-point targeting, similar to methods used in previous works \cite{Dehghani-23, Kim-23}. 
We selected two virtual target layer indices, at $p=40\%$ and $p=60\%$, where the latter was selected to simulate injections closer to the RPE. Furthermore, we empirically select two maximum velocities at 0.3 mm/s and 0.4 mm/s based on preliminary experiments to ensure the needle can puncture the ILM. The factor $\alpha$ was set to 0.1 for all experiments.
Additionally, we incorporated a syringe pump with an infusion rate of $1 ml/min$ and a target volume of $0.1 ml$ to simulate subretinal injections.
We performed five insertions for each velocity, target depth and method combination.

The selection of the target point using fixed-point targeting was done using the following approach:
First, the needle was placed at the insertion point on the ILM. Then, an OCT volume was acquired, and the B-scan slice closest to the needle center was manually selected. 
Afterward, the pixel coordinates of the target point along the needle axis located at $L_{\text{target}}$ were identified. 
These pixel coordinates were transformed, with corrections to the axial spacing using the method described in \cite{Dehghani-23}. 
Finally, the robot was moved along the needle axis at a constant velocity until the fixed target point was reached. 
It is important to note that in this variant, the target point was only calculated before and was not updated during the insertion, not compensating for tissue deformations.

\section{RESULTS}
\subsection{Relative Depth Calculation Accuracy}
In Table \ref{tab:rel-dep-accuracy-res} 
we present the average error of the proposed method to identify the depth of the needle tip, the ILM and the RPE at the A-scans that coincide with the needle tip.
This error is converted into \textmu m from manually labelled B$^5$-scans based on their voxel spacing. 
The average retinal thickness on the A-scan corresponding to the needle tip is also presented in the results.

\begin{table}[]
\centering
\caption{Results of our needle tip and layer processing methods. All errors in \textmu m}
\label{tab:rel-dep-accuracy-res}
\begin{tabular}{lllll}
\hline
\textbf{Phase}   & \textbf{\begin{tabular}[c]{@{}l@{}}Avg. needle \\ depth error \end{tabular}} & \textbf{\begin{tabular}[c]{@{}l@{}}Avg. ILM\\ depth error\end{tabular}} & \textbf{\begin{tabular}[c]{@{}l@{}}Avg. RPE \\ depth error\end{tabular}} & \textbf{\begin{tabular}[c]{@{}l@{}}Avg. retina \\ thickness\end{tabular}} \\ \hline
\textbf{Inside}  & 13.4 $\pm$ 17.3                                                             & 8.2 $\pm$ 4.7                                                           & 21.4 $\pm$ 8.1                                                           & 448 $\pm$ 54                                                              \\
\textbf{Outside} & 8.2 $\pm$ 12.8                                                              & 24.1 $\pm$ 19.1                                                         & 45.8 $\pm$ 76.9                                                          & 409 $\pm$ 75                                                              \\
\textbf{Border}  & 9.9 $\pm$ 9.9                                                               & 24.1 $\pm$ 18                                                           & 13.2 $\pm$ 5.1                                                           & 371 $\pm$ 29                                                              \\
\textbf{Total}   & 10.7 $\pm$ 13.7                                                             & 18.2 $\pm$ 16.3                                                         & 26.5 $\pm$ 44                                                            & 412 $\pm$ 63                                                              \\ \hline
\end{tabular}
\end{table}

\begin{table*}[]
\caption{Results of comparing our proposed method with a fixed point targeting method}
\label{tab:comparison-res}
\resizebox{\textwidth}{!}{%
\begin{tabular}{lcccccccc}
\hline
\textbf{}                         & \multicolumn{2}{c}{\textbf{Final error (\textmu m)}}                                                                   & \multicolumn{2}{c}{\textbf{Final error (\%)}}                                                                                         & \multicolumn{2}{c}{\textbf{Retina thickness (\textmu m)}}                                                               & \multicolumn{2}{c}{\textbf{Bleb}}                                                                                                      \\ \hline
\textbf{Target depth}             & \begin{tabular}[c]{@{}c@{}}Fixed Point\\ Targeting\end{tabular} & \textbf{\begin{tabular}[c]{@{}c@{}}Proposed \\ Method\end{tabular}} & \begin{tabular}[c]{@{}c@{}}Fixed Point\\ Targeting\end{tabular} & \textbf{\begin{tabular}[c]{@{}c@{}}Proposed \\ Method\end{tabular}} & \begin{tabular}[c]{@{}c@{}}Fixed Point \\ Targeting\end{tabular} & \textbf{\begin{tabular}[c]{@{}c@{}}Proposed \\ Method\end{tabular}} & \begin{tabular}[c]{@{}c@{}}Fixed Point \\ Targeting\end{tabular} & \textbf{\begin{tabular}[c]{@{}c@{}}Proposed \\ Method\end{tabular}} \\ \hline
\textbf{0.4 mm/s needle velocity} & \multicolumn{1}{l}{}                                            & \multicolumn{1}{l}{\textbf{}}                                       & \multicolumn{1}{l}{}                                            & \multicolumn{1}{l}{\textbf{}}                                       & \multicolumn{1}{l}{}                                             & \multicolumn{1}{l}{\textbf{}}                                       & \multicolumn{1}{l}{}                                             & \multicolumn{1}{l}{\textbf{}}                                       \\ \hline
\textbf{40\%}                     & 56 $\pm$ 27                                                     & \textbf{9 $\pm$ 4}                                                  & 47 $\pm$ 40                                                     & \textbf{6 $\pm$ 4}                                                  & 357 $\pm$ 118                                                    & \textbf{409 $\pm$ 123}                                              & 3/5                                                              & \textbf{4/5}                                                        \\
\textbf{60\%}                     & 140 $\pm$ 60                                                    & \textbf{30 $\pm$ 11}                                                & 77 $\pm$ 39                                                     & \textbf{10 $\pm$ 4}                                                 & 317 $\pm$ 60                                                     & \textbf{503 $\pm$ 45}                                               & 1/5                                                              & \textbf{5/5}                                                        \\ \hline
\textbf{0.3 mm/s needle velocity} & \multicolumn{1}{l}{}                                            & \multicolumn{1}{l}{\textbf{}}                                       & \multicolumn{1}{l}{}                                            & \multicolumn{1}{l}{\textbf{}}                                       & \multicolumn{1}{l}{}                                             & \multicolumn{1}{l}{\textbf{}}                                       & \multicolumn{1}{l}{}                                             & \multicolumn{1}{l}{\textbf{}}                                       \\ \hline
\textbf{40\%}                     & 81 $\pm$ 48                                                     & \textbf{19 $\pm$ 11}                                                & 55 $\pm$ 33                                                     & \textbf{10 $\pm$ 6}                                                 & 363 $\pm$ 39                                                     & \textbf{475 $\pm$ 71}                                               & 1/5                                                              & \textbf{4/5}                                                        \\
\textbf{60\%}                     & 125 $\pm$ 60                                                    & \textbf{11 $\pm$ 9}                                                 & 64 $\pm$ 36                                                     & \textbf{4 $\pm$ 3}                                                  & 342 $\pm$ 63                                                     & \textbf{496 $\pm$ 47}                                               & 2/5                                                              & \textbf{5/5}                                                        \\ \hline
\textbf{Total}                    & 100 $\pm$ 58                                                    & \textbf{17 $\pm$ 12}                                                & 61 $\pm$ 36                                                     & \textbf{7.9 $\pm$ 5}                                                & 345 $\pm$ 72                                                     & \textbf{471 $\pm$ 81}                                               & 7/20                                                             & \textbf{18/20}                                                      \\ \hline
\end{tabular}%
}
\end{table*}

\subsection{Comparison of Targeting Methods}
In Table \ref{tab:comparison-res}, we show the results of our experiments guiding a needle into the subretinal space using a fixed target point and our virtual target layer methods, respectively.
We evaluate the final needle tip positioning, based on manually labeled B$^5$-scans, to the virtual target layer. 
The results are converted to \textmu m using the voxel spacing of the volumes.

\begin{figure}[h]
    \centering
    \includegraphics[width=\linewidth]{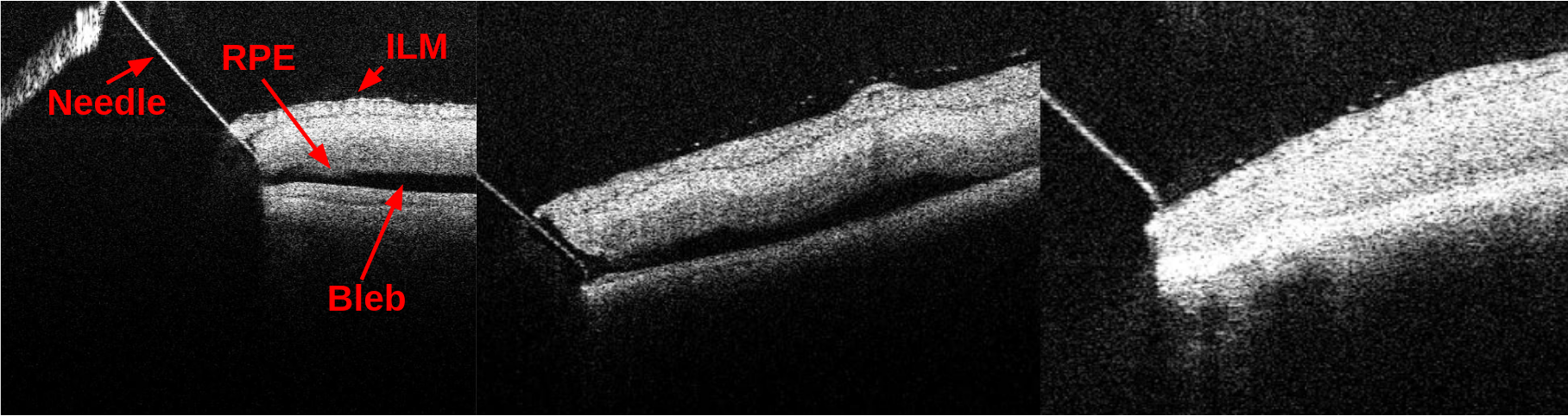}
    \caption{Examples of our successful (left and middle) and unsuccessful (right) bleb definitions, In successful cases the injected liquid causes large retinal detachments in post-mortem eyes rather than the typical bleb formation seen in \textit{in-vivo} procedures. In unsuccessful cases although the needle starts in the retina, the injection pushes the retina away causing no liquid accumulation under the retina.}
    \label{fig:bleb-examples}
\end{figure}

Due to the use of \textit{ex-vivo} open-sky porcine eyes, creating high-quality blebs is challenging compared to intact or \textit{in-vivo} eyes due to an increased risk of retinal detachment. 
Therefore, for this experiment, we define a successful bleb generation as the event of subretinal fluid accumulation (Fig. \ref{fig:bleb-examples}). 
In unsuccessful insertions, the injection displaces the retina and introduces fluid into the vitreous without fluid accumulation below the retina.

\subsection{Inference Time Analysis}
Using an NVIDIA GeForce RTX 2080 Ti GPU, the inference time for our segmentation network is 20 ms to process a full B$^{5}$-scan and the following processing steps take 47 ms. The acquisition time for a B$^{5}$-scan is 115 ms. Our analysis revealed that the current bottleneck in the system is the data acquisition process, which limits the update rate to approximately 9 Hz. The combined processing time for segmentation and relative depth calculation (67 ms) is shorter than the acquisition time, ensuring that necessary computations can be completed before the next B$^{5}$-scan is acquired. 

\section{DISCUSSION AND FUTURE WORK}
In our experiments, we evaluated the accuracy of our method in determining the needle tip, ILM and RPE positions in B$^{5}$-scans and compared our robot control strategy to previously used fixed-point targeting methods.

\begin{figure}
    \centering
    \includegraphics[width=0.87\linewidth]{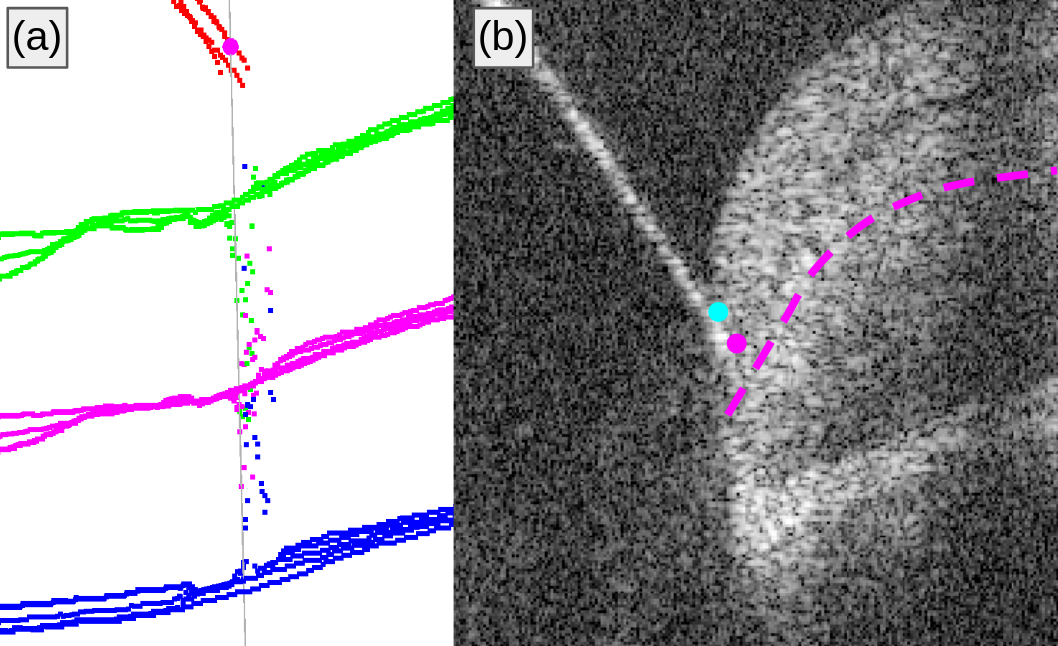}
    \caption{(a) Example of incorrect layer inpainting (ILM green, RPE blue) and its effects on virtual target layer (pink). (b) Significantly deformed ILM during needle insertion and the correct needle tip (pink) and incorrect needle tip (cyan). Due to its larger distance to the target layer, the incorrect tip point would cause the needle to be inserted damaging the RPE more, while the correct tip point would stop the insertion.}
    \label{fig:common-errors}
\end{figure}

One notable observation was the high variance in calculating the position of the ILM and RPE.
We believe these errors can be attributed to artifacts generated by the inpainting procedure, which aims to reconstruct blocked tissue areas from neighboring tissue information, created artifacts in cases where large regions with no or only limited neighbor information were inpainted (Fig.~\ref{fig:common-errors}(a)).
The RPE was affected more often by this effect than the ILM. 

Increased errors can also be seen in cases where the ILM, consequently the virtual target layer, is significantly deformed (Fig. \ref{fig:common-errors}(b)). Because of its steep angle, small errors in needle tip position could lead to large discrepancies in the robot control.


Furthermore, we noted an increase in needle error as the needle was inserted deeper into the tissue due to less distinct visual boundaries between the instrument and the surrounding tissue, posing more significant challenges for accurate depth calculation.

Our second experimental evaluation compared our method with the fixed-point targeting method used in \cite{Dehghani-23}. 
These experiments highlight the effects of tissue deformation and underscore the necessity for a deformation-aware control strategy for successful subretinal injection.

By comparing the needle insertion to two different virtual target layers, we observed that the fixed-point targeting pushed the retina down without reaching the relocated target, in some cases failing to penetrate the ILM altogether.
Our method was able to overcome these limitations by continuously monitoring the needle's position in relation to the virtual target layer, ensuring its correct placement relative to the RPE.
Our final axial error, 17 \textmu m, is acceptable based on the anatomical target area of 25-30 \textmu m \cite{Karampelas2013-ch}.
For reference \cite{Dehghani-23} achieved an average 32 \textmu m Euclidean error in reaching the target.


Based on the successful bleb creation frequency, our method proved superior in achieving accurate anatomical targeting by reliably producing blebs.
The comparison method, on the other hand, often resulted in fluid being injected into the vitreous.
We believe this is due to the needle's ability to puncture the ILM and position itself more effectively within the retina. 
This suggests that accurate anatomical targeting is critical for successful bleb creation and fluid delivery.
Additionally, our method was slower, taking on average 2.8 seconds per needle insertion. 

One notable issue is the \textit{bounce back} effect, previously described in \cite{pannek}. 
This phenomenon occurs when the tissue recoils closer to the start position following needle penetration, causing the needle to advance relative to the ILM position but not the RPE position. 
Due to system latencies, our method may not always react quickly enough to identify this effect, leading to overshooting the target and risking damage to the RPE layer.

Discussions with experts revealed that this effect is relatively common in live tissue. 
We believe this effect is made more quantitatively significant due to post-mortem changes, in the tissue during the decomposition process.
One physical example of such changes is the swelling of the retinal tissue, resulting in the relatively large average retinal thickness in our results.
To better understand the impact of the \textit{bounce-back} effect and assess its relevance to our method, further studies should be conducted using live eyes.

\section{CONCLUSION}
In this paper, we presented a pipeline for real-time subretinal needle insertion control for autonomous robotic subretinal injection by monitoring the needle tip position relative to a virtual target layer, defined at a relative depth between the ILM and RPE. 
Our approach is enabled by the continuous acquisition of B$^{5}$-scans, densely sampled 3D OCT volumes, to achieve real-time imaging and processing performance with update rates of 9 Hz.


These results demonstrate a method for more reliable needle placement and enhances the success of subretinal injections, potentially improving needle insertions for future autonomous robotic subretinal injections. 






\newpage
\bibliographystyle{IEEEtran}
\bibliography{root}
\end{document}